\documentclass[conference]{IEEEtran}
\ifCLASSINFOpdf
\else
\fi

\usepackage[pdftex]{graphicx}
\usepackage{caption}
\usepackage{amsmath}
\usepackage{amssymb}
\usepackage{amstext}
\usepackage{amsfonts}
\usepackage{url}
\usepackage{bbm}
\usepackage{color}
\usepackage[table,xcdraw]{xcolor}

\begin{document}
%
\title{ICDAR2017 Competition on Reading Chinese\\Text in the Wild (RCTW-17)}

\author{\IEEEauthorblockN{Baoguang Shi$^{1}$, Cong Yao$^{2}$, Minghui Liao$^{1}$, Mingkun Yang$^{1}$,\\Pei Xu$^{1}$, Linyan Cui$^{1}$, Serge Belongie$^{3}$, Shijian Lu$^{4}$, Xiang Bai$^{1}$}
\IEEEauthorblockA{$^{1}$School of EIC, Huazhong University of Science and Technology, China\\
$^{2}$Megvii Technology Inc., China\\
$^{3}$Department of Computer Science, Cornell University \& Cornell Tech, USA\\
$^{4}$School of Computer Science and Engineering, Nanyang Technological University, Singapore}
}


%


\maketitle

\begin{abstract}
Chinese is the most widely used language in the world. Algorithms that read Chinese text in natural images facilitate applications of various kinds. Despite the large potential value, datasets and competitions in the past primarily focus on English, which bares very different characteristics than Chinese. This report introduces RCTW, a new competition that focuses on Chinese text reading. The competition features a large-scale dataset with 12,263 annotated images. Two tasks, namely text localization and end-to-end recognition, are set up. The competition took place from January 20 to May 31, 2017. 23 valid submissions were received from 19 teams. This report includes dataset description, task definitions, evaluation protocols, and results summaries and analysis. Through this competition, we call for more future research on the Chinese text reading problem.
The official website for the competition is \url{http://rctw.vlrlab.net}
\end{abstract}


%
\IEEEpeerreviewmaketitle

\section{Introduction}

Text in natural images is an important information source. Algorithms that reads text in natural images facilitate a lot of real world applications, such as geo-location and fine-grained image classification. Driven by the increasing amount of image data and the popularity of mobile devices, scene text reading has been receiving much attention from both the academia and the industry.

It is widely recognized that large-scale, well-annotated datasets are crucial to the success of computer vision algorithms and systems. In the past few years, many scene text datasets have been collected for research and product. Datasets such as ICDAR 2013~\cite{icdar/KaratzasSUIBMMMAH13}, SVT~\cite{eccv/WangB10}, MSRA-TD500~\cite{cvpr/YaoBLMT12}, and ICDAR 2015 Incidental Text~\cite{icdar/KaratzasGNGBIMN15} have gain much popularity in this field, and have become standard benchmarks for algorithms and systems.

Despite the plenty of publicly available data, most focuses on English text. Chinese text reading has been less studied in this field. As Chinese is the most widely used language around the world, Chinese text reading has a large potential practical value. Moreover, Chinese text has different characteristics compared with English text: Chinese has a much larger character set than English; Chinese words are not separated by blank spaces; Many Chinese characters are made of multiple non-connected parts. Because of these characteristics, reading Chinese in the wild is a unique problem.

Realizing its potential value, we propose a new competition for Chinese text reading. This competition features a new image database of Chinese scene text. The dataset has more than 12,000 images with detailed annotations, including the location and transcription of every text instance. We set up two tasks for this competition, namely text localization and end-to-end recognition.

The competition was held from January till the end of April 2017. It received a good deal of attention from the community. 59 teams registered for participation. Among them, 19 submitted their results. In this report, we present their evaluation results and analysis.

Through the new dataset and tasks, we aim to call for more future research and development efforts on the problems of Chinese text reading.

\section{Dataset and Annotations}

\begin{figure*}
\centering
\includegraphics[width=1.0\linewidth]{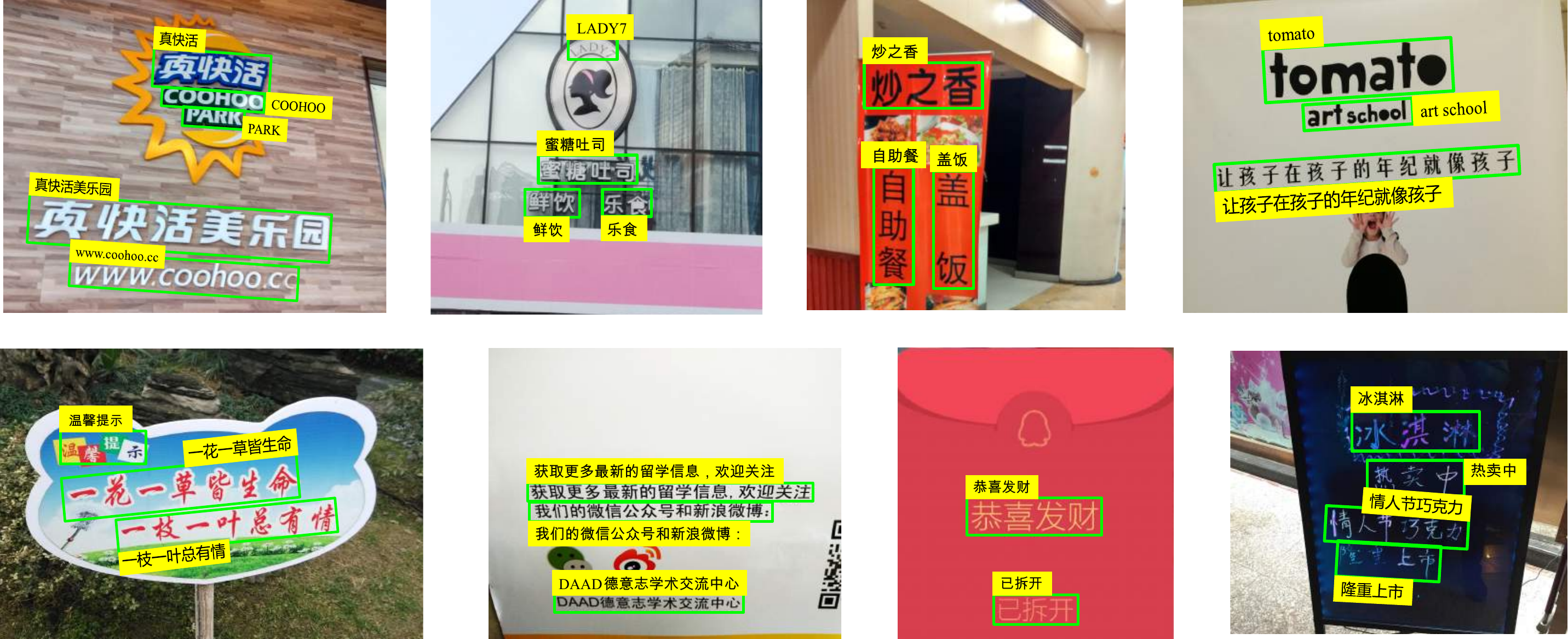}
\caption{Example images and annotations of the CTW-12k dataset.}
\label{fig:dataset-examples}
\end{figure*}

The dataset is named CTW-12k, as it comprises 12,263 images of Chinese Text in the Wild. Most images are natural images collected by ourselves using phone cameras. Others are ‘digital born’. They are mostly screen-shots taken from smart phones or personal computers. All images contain at least one line of Chinese text.

We manually annotated all images by drawing polygons to surround every text line, using a tool we created. A polygon comprises four points arranged in the clockwise order, starting from the top-left point. After that, each text line is annotated with its transcription encoded as UTF-8 strings. Text lines that are illegible are marked by a “difficult” flag. In our dataset, we annotate text at the level of text lines. Words are not separated, since Chinese words do not have blank spaces between each other.

Figure~\ref{fig:dataset-examples} shows example images and their annotations. Note the diversities in image sources, text fonts, layouts, and languages (Chinese and English).

The dataset is split into two subsets. The training and validation (`trainval') set consists of 8,034 images. Images and annotations of this dataset were released during the competition. The test set comprises 4,229 images. The test set images were released one week before the submission deadline.

\section{Challenge Tasks}

We set up two competition tasks: text localization and end-to-end recognition. Unlike many former competitions, we did not set up a cropped text recognition task, for we think that the recognition performance is better evaluated under the end-to-end recognition setting.

\subsection{Task 1 - Text Localization}

Text localization (aka. detection) is a conventional competition task. The objective of this task is to localize text instances in images by polygons with four points. A score should also be provided for every polygon, indicating the detection confidence.


\begin{figure}
\centering
\includegraphics[width=0.7\linewidth]{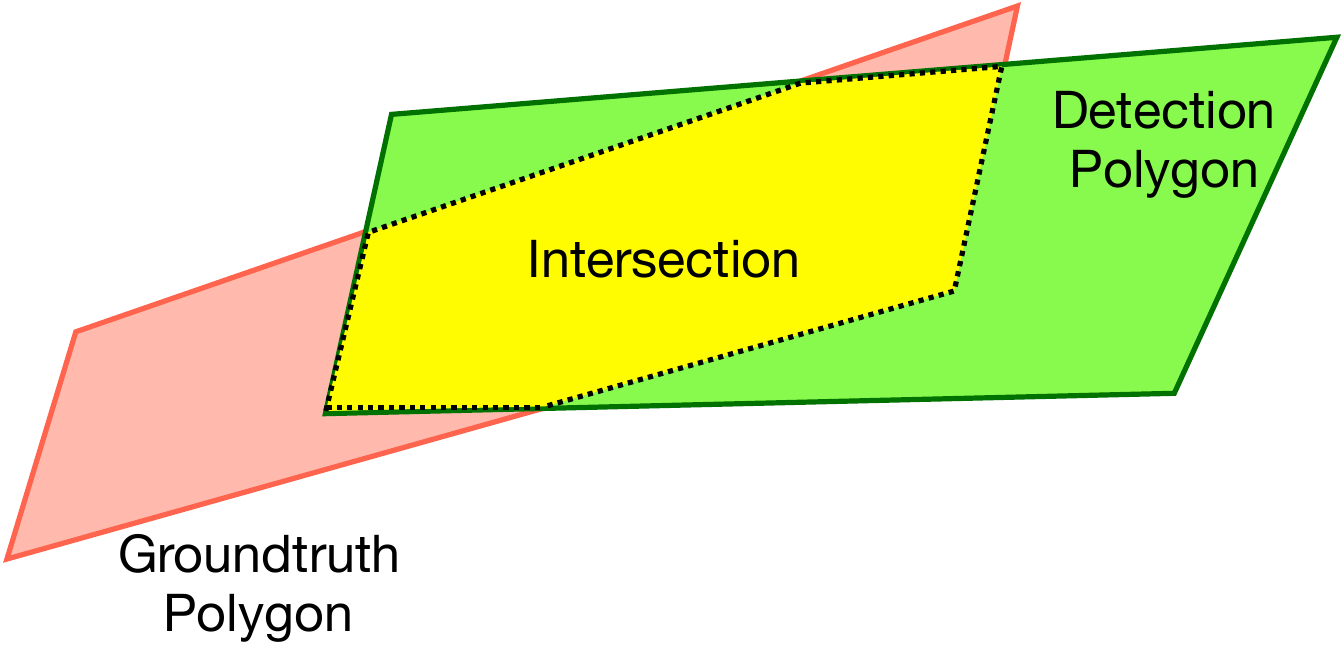}
\caption{Intersection-over-Union (IoU) of two polygons. The red and green polygons are groundtruth and detection polygons respectively. The yellow area is their intersection. Union area is defined as the sum of the two polygon areas minus their intersection area. IoU is the ratio between intersection and union areas.}
\label{fig:iou}
\end{figure}

For this task, the evaluation protocol follows that of PASCAL VOC~\cite{ijcv/EveringhamGWWZ10}, which adopts mean Average Precision (mAP) as the primary metric. Basically, mAP is the normalized area below precision-recall (PR) curve, averaged across object categories. Since text is the only foreground object category in our competition, the metric is AP.

The original AP is defined on axis-aligned bounding boxes, while in this competition text is localized by polygons. Considering this, we calculate intersection-over-union (IoU) on polygons rather than rectangles. The IoU is illustrated in Figure~\ref{fig:iou}. We implement the calculation of IoU using the Shapely package\footnote{\url{https://pypi.python.org/pypi/Shapely}}.

A detection is matched to a groundtruth and marked as true positive if 1) IoU is over 0.5 and 2) the groundtruth is not matched to another detection. When multiple detections are matched to the same groundtruth, we only pick the one with the highest IoU.

Former competitions commonly used F-score, the harmonic mean of precision and recall, as their primary metric. F-score is sensitive to the trade-off between precision and recall. Therefore, careful tuning is often needed to get the best combination of precision and recall. On the other hand, AP is invariant to that trade-off. AP, on the other hand, calculates a measure by iterating over all points of precisions and recalls. Thus, it is invariant to the trade-off between precision and recall. By calculating AP, we also get precision-recall curves (PR-curve) as the byproducts. PR-curve provides a comprehensive view on detector performance.

We take AP the primary metric and rank submissions according to it. For compatibility with former competitions, we also calculate a maximum F-measure score for every submission. It is found across all points of PR curve. Both scores are listed in the results.

\subsection{Task 2 - End-to-End Recognition}

The aim of this task is to both localize and recognize text. Participants were asked to submit detection results along with recognition results. The submission format is similar to that of Task 1. Detection scores are replaced by recognized text encoded as UTF-8 strings.

Method performance was evaluated by the edit distances between recognized text and groundtruth text. The evaluation process consists of two steps. First, every detection is matched to either 1) a groundtruth polygon that has the maximum polygon IoU, or 2) `None' if none of the groundtruth polygons has over 0.5 IoU with the detection. If multiple detected polygons are matched to the same groundtruth polygon, only the one with the maximum IoU will be kept and rest are matched to `None'.
After that, we calculate the edit distances between all matching pairs.
If a detection is matched to `None', edit distance is calculated between the recognized text and an empty string.
The edit distances are summed and divided by the number of test images.
Method is scored by the average edit distance (AED). Lower AED means better performance.

This metric concerns both detection and recognition performance. If detection fails, either by generating a false positive or false negative, the metric puts a penalty equal to the length of groundtruth text.

For compatibility with other competitions, we also calculate a normalized measure. We first calculate the normalized edit distance (NED) as $\mathrm{NED(s_1, s_2)} = \mathrm{edit\_dist}(s_1, s_2) / \max(l_1, l_2)$ where $s_1$ and $s_2$ are the text strings of a matching pair and $l_1$, $l_2$ are their text lengths. Then, we calculate the measure using $1 - \sum_{i=1}^{n} \mathrm{NED(s_{i1}, s_{i2})} / n$, where $n$ is the number of matching pairs.

During the evaluation, we exclude text instances that are marked by the ``difficult'' flag. Detections that are matched to such instances increase zero edit distance. However, we do not exclude such instances in Task 1. Although such instances are not recognizable, humans mostly have no difficulties localizing them in images.

We did not adopt the more popular evaluation protocol proposed by Wang \emph{et al.}~\cite{eccv/WangB10}, which considers detection as a match if IoU exceeds 0.5 \emph{and} recognition matches exactly with groundtruth. Under that protocol, longer text instances and shorter text instances contribute equally to the final score.
In contrast, under our protocol, longer text instances contribute larger penalties.
Therefore, our protocol demands better long text detection and recognition performance, which is favorable in practical systems.

\section{Organization}

The competition started on January 20, 2017, when we make our website public. The website provides competition info, dataset download links (not released at the beginning), a registration page, and a submission page. The training/validation set was released on February 15; The test dataset was released on April 15, two weeks before the submission deadline. Before releasing the test dataset, we revised the training/validation set once to fixed some annotation errors. The submission portal was opened on April 17 and closed at 11:59 PM PST, April 30.

We received all together 59 registrations. Most of the registrations come from Universities, research institutes, and tech companies in China. We also received some registrations from Universities in United States, United Kingdom, and Australia.

Teams submitted their results by sending us an email with their compressed results attached. By the submission deadline, we received submissions from 19 teams. All of the teams submitted results for Task 1. Four of them also submitted results for Task 2. Due to the complexity of Task 2, the much less number of submissions was expected. Some teams submitted multiple results for the same task. We only took their last submissions, discarding the rest.

\section{Submissions and Results}

\begin{figure}
\centering
\includegraphics[width=0.8\linewidth]{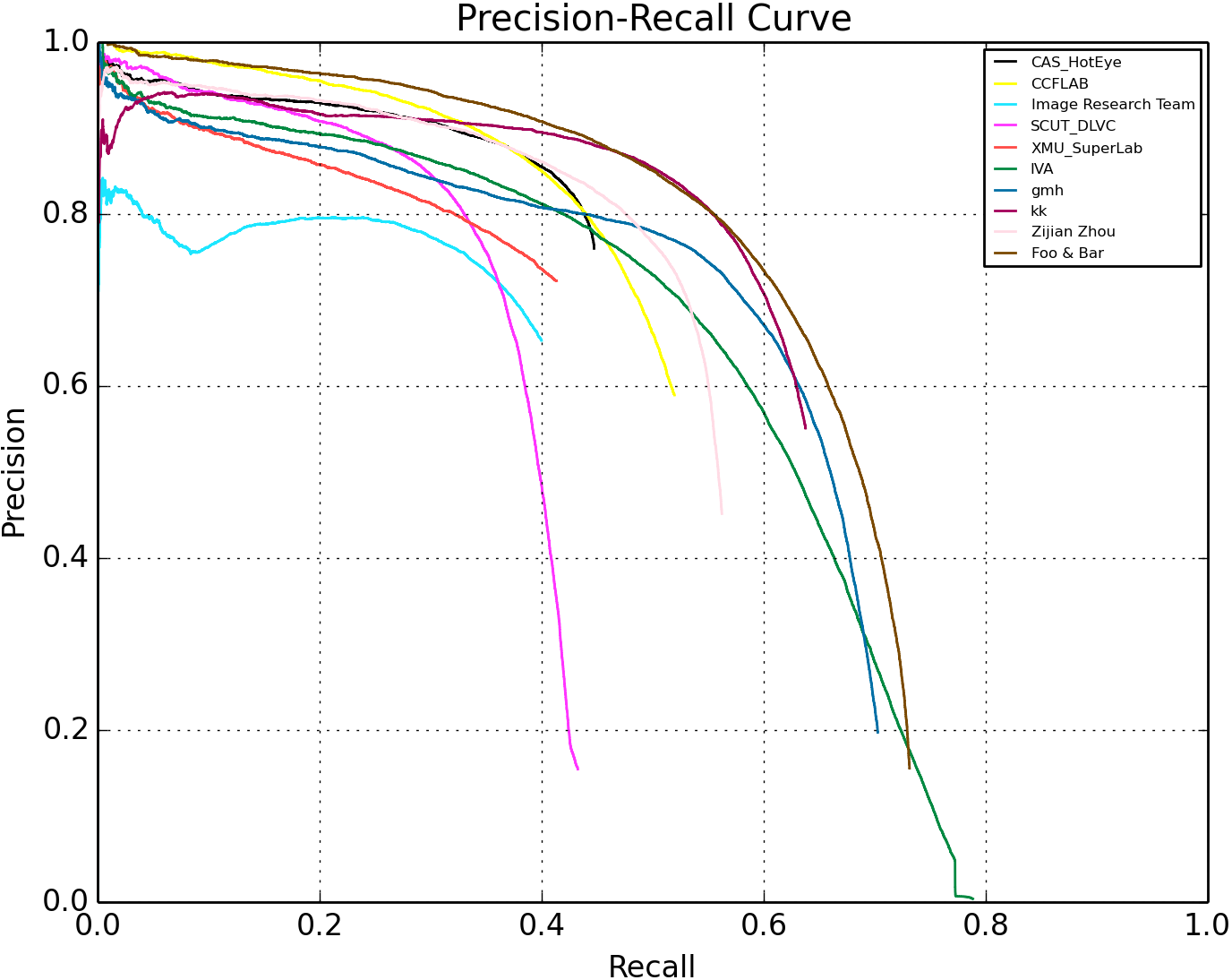}
\caption{Summary of PR curves of the top-10 submissions. Each curve represents a team. Viewed in color.}
\label{fig:pr-summary}
\end{figure}

\definecolor{Gray}{gray}{0.9}
\begin{table*}[t]
\centering
\caption{Results summary for the top-10 submissions of Task 1. A-rank stands for Average Precision rank; F-rank stand for Maximum F-measure rank.}
\label{tbl:results-task1}
\renewcommand{\arraystretch}{1.7}
\begin{tabular}{p{2cm}p{3cm}llllllp{3cm}}
\hline
\rowcolor{Gray}
\textbf{Team Name} & \textbf{Team Member} & \textbf{AP} & \textbf{A-Rank} & \textbf{F-meausre} & \textbf{F-Rank} & \textbf{Precision} & \textbf{Recall} & \textbf{Institute} \\ \hline
Foo \& Bar & Zheqi He,Yongtao Wang & 0.623447 & 1 & 0.661054 & 1 & 0.743876 & 0.594827 & Peking University \\
\rowcolor{Gray}
NLPR\_PAL & Wenhao He, Fei Yin, Da-Han Wang, Cheng-Lin Liu & 0.560427 & 2 & 0.657598 & 2 & 0.771675 & 0.572905 & NLPR,CASIA \\
gmh & Minghao Guo & 0.555034 & 3 & 0.636024 & 3 & 0.706367 & 0.578422 & Tsinghua University \\
\rowcolor{Gray}
SCUT\_MBCNN & Jinrong Li, Zijian Zhou, Shuangping Huang & 0.494374 & 5 & 0.608396 & 4 & 0.736135 & 0.518434 & South China University of Technology \\
IVA & Hao Ye, Yingbin Zheng, Weiyuan Shao, Hong Wang & 0.554721 & 4 & 0.601721 & 5 & 0.661029 & 0.552179 & Shanghai Advanced Research Institute,CAS \\
\rowcolor{Gray}
CCFLAB & Dai Yuchen, Huang Zheng, Gao Yuting & 0.468102 & 6 & 0.576006 & 6 & 0.740618 & 0.471261 & Shanghai Jiao Tong University \\
CAS\_HotEye & Wu Dao, Daipeng Wen & 0.408581 & 7 & 0.56697 & 7 & 0.791453 & 0.441691 & Instutute of Information Engineering,CAS \\
\rowcolor{Gray}
XMU\_SuperLab & Xiaodong Yang, Li Lin, Yan Zhang, Jinyan Liu, Weiran Li, Bin Jin & 0.351821 & 9 & 0.525778 & 8 & 0.722228 & 0.413346 & Xiamen University \\
Image Research Team & Long Ma, Lulu Xu, Shenghui Xu & 0.312182 & 10 & 0.496194 & 9 & 0.654381 & 0.399597 & Sogou Inc. \\
\rowcolor{Gray}
SCUT\_DLVC & Lianwen Jin, Yuliang Liu, Zenghui Sun, Canjie Luo, Zhaohai Li, Lele Xie, Fan Yang & 0.360008 & 8 & 0.481663 & 10 & 0.705839 & 0.36556 & South China University of Technology \\
\hline
Baseline & Minghui Liao & 0.359432 & N/A & 0.527837 & N/A & 0.760318 & 0.404385 & Mclab, Huazhong University of Science and Technology \\ \hline
\end{tabular}
\end{table*}

\begin{table*}[t]
\centering
\caption{Results summary for submissions of Task 2. AED stands for Average Edit Distance. Normalized is the normalized measure.}
\label{tbl:results-task2}
\renewcommand{\arraystretch}{1.7}
\begin{tabular}{lp{4cm}lllp{5cm}}
\hline
\rowcolor{Gray}
\textbf{Team Name} & \textbf{Team Member} & \textbf{AED} & \textbf{AED-Rank} & \textbf{Normalized} & \textbf{Institute} \\
\hline
NLPR\_PAL & Yan-Fei Lv, Wenhao He, Fei Yin, Cheng-Lin Liu & 20.21967368 & 1 & 0.3201 & NLPR,CASIA \\
\rowcolor{Gray}
SCUT\_DLVC & Lianwen Jin, Yuliang Liu, Zenghui Sun, Canjie Luo, Zhaohai Li, Lele Xie, Fan Yang & 28.3078742 & 2 & 0.2374 & South China University of Technology \\
CCFLAB & Dai Yuchen, Huang Zheng, Gao Yuting & 32.129818 & 3 & 0.2143 & Shanghai Jiao Tong University \\
\rowcolor{Gray}
Image Research Team & Long Ma, Lulu Xu, Shenghui Xu & 35.28943013 & 4 & 0.1577 & Sogou Inc. \\
\hline
Baseline & Mingkun Yang & 25.62260582 & N/A & 0.2412 & Mclab, Huazhong University of Science and Technology \\
\hline
\end{tabular}
\end{table*}

Our evaluation program was implemented in Python\footnote{Source code is available at \url{https://github.com/bgshih/rctw17}}. We run the program to evaluate all submissions after the deadline. Table~\ref{tbl:results-task1} summarizes the top-10 submissions of Task 1. We include both AP and Maximum F-measure in the results table. Methods were ranked by their AP numbers. Ranking by F-measure produces similar ranking results. For a more detailed comparison, we visualized the PR curves of all methods in Figure~\ref{fig:pr-summary}.

\begin{figure}[!t]
\centering
\includegraphics[width=1.0\linewidth]{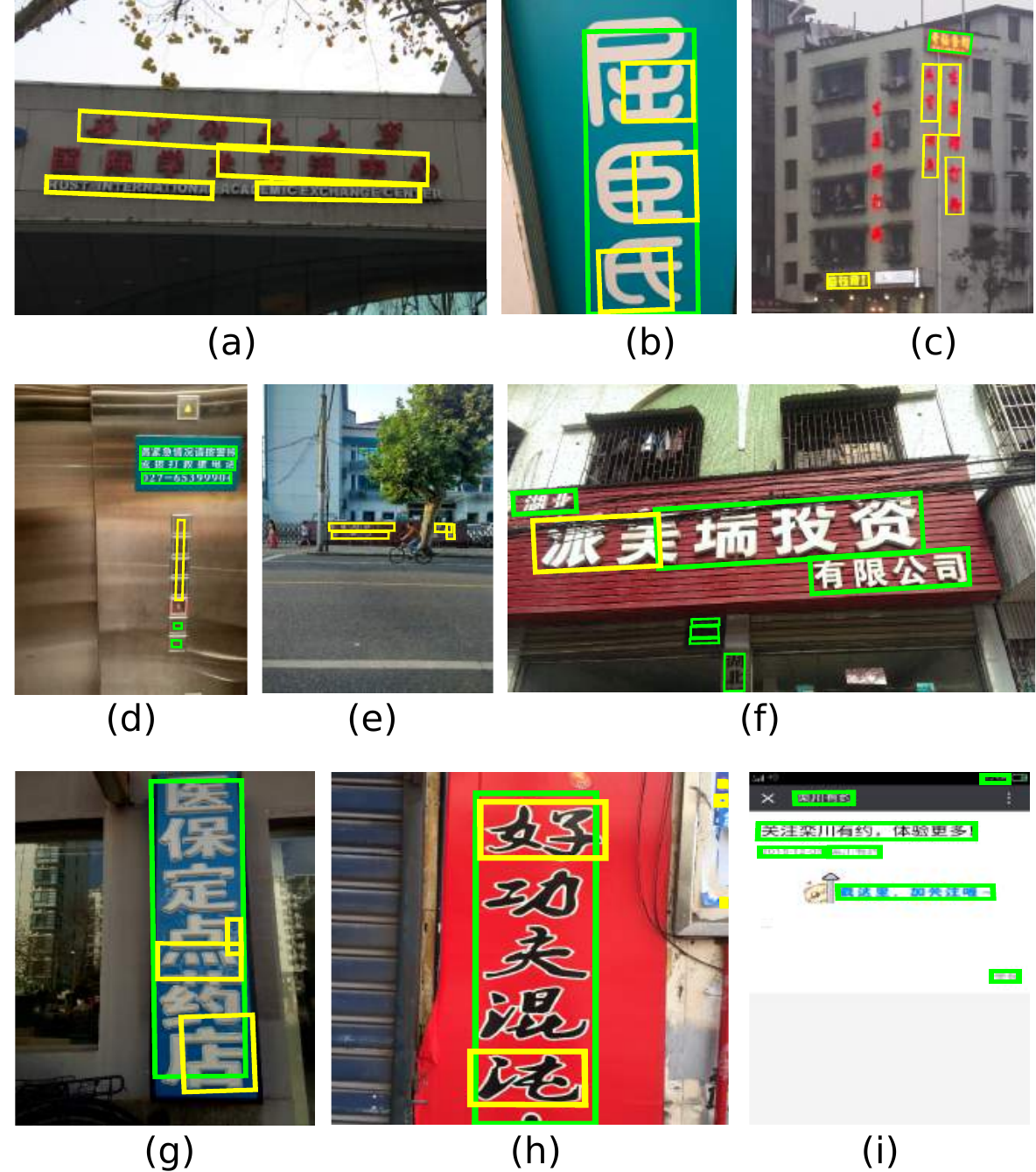}
\caption{Example detections from the submissions. Green polygons are correctly detected. Yellow ones are false detections.}
\label{fig:task1-examples}
\end{figure}

\subsection{Top 3 submissions for Task 1}

\textbf{1. ``Foo\&Bar'' (Peking University)} The method used Faster R-CNN~\cite{pami/RenHG017} based method and replace bounding box regression layer to their quadrangular layer, which can get 4 points of quadrangle. They used res101 and the model was pretrained in ImageNet.

\textbf{2. ``NLPR\_PAL'' (CASIA)} The method used Deep Direct Regression Network~\cite{corr/HeZYL17}, which produces text bounding boxes or text line segments (for too long lines). Segments are grouped by a line grouping method into text lines. Their training dataset contains both the training set provided by RCTW17 and 7000 images selected by themselves.

\textbf{3. ``gmh'' (Tsinghua University)} A CNN based method.

\subsection{Top 3 submissions for Task 2}

\textbf{1. ``NLPR\_PAL'' (CASIA)} The system is based on sliding convolutional character models, which are learned end to end on text line images  labeled with text transcripts. The character classifier outputs on the sliding windows are normalized and decoded  with refined Beam Search based algorithm. The system is purely trained with synthetic dataset in 7,356 classes  (Chinese characters and symbols).

\textbf{2. ``SCUT\_DLVC'' (South China University of Technology)} The detection part consists of two stages. The first stage generate rough text candidate quadrilaterals; the second stage refine the candidates. The recognition part combines multi-scale CNN, bi-directional LSTM, and CTC for sequential text recognition. They proposed a perspective transform method to utilize the detected quadrilaterals for better recognition. They obtained extra training data by synthesizing in a similar manner as~\cite{cvpr/GuptaVZ16}.

\textbf{3. ``CCFLAB'' (Shanghai Jiao Tong University)} Their method is based on Faster RCNN, and they use ResNet-101 as the backbone network. They use ImageNet pre-trained model to initialize the whole architecture. To improve the performance for detecting small text regions, they use fused feature map from a top-down order, followed by ROI pooling to get fixed-size feature on which Fast-RCNN branch should do final detections. During training, they only counter on the provided dataset, without using extra data.

\subsection{Baseline submissions}

For reference, we submitted a baseline method to Task 1 and Task 2 respectively. The methods were implemented by ourselves. Their results are shown in Table~\ref{tbl:results-task1} and Table~\ref{tbl:results-task2}.

For Task 1, the text detection method is based on SegLink~\cite{cvpr/ShiBB17}. A fully convolutional network is applied to predict segments and links at multiple scales. Segments connected by links are combined into text lines. We use the same setting as~\cite{cvpr/ShiBB17}. Only the training data changed.

For Task 2. The detection part is the same as that for Task 1. We took the Chinese text line recognition as a sequence recognition task. A modified version CRNN~\cite{corr/ShiBY15} is adopted, which uses the convolutional layers to extract features, the Bi-LSTM layers to capture the spatial context, and CTC for transcription without char-level annotation in advance. With the detected results in task1, the text lines are cropped and classified into horizontal or vertical by the length-width ratio. Then, different models are applied respectively. We used a large synthetic dataset with a Chinese lexicon to pre-train our model.

\begin{figure}[!t]
\centering
\includegraphics[width=1.0\linewidth]{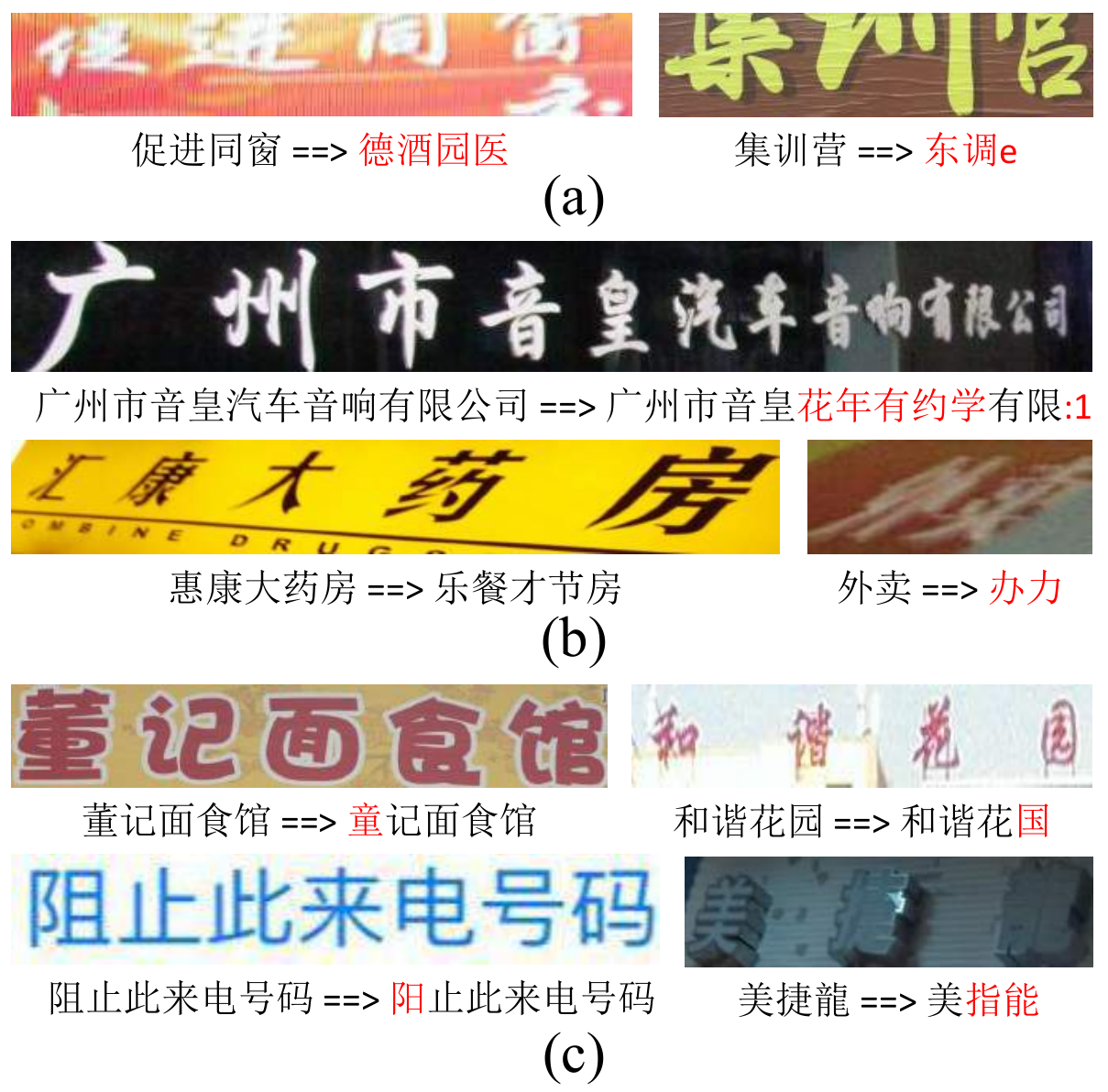}
\caption{Examples of recognition. Red characters are recognized wrongly.}
\label{fig:task2-examples}
\end{figure}

\section{Analysis}

To have a better understanding on performance, we visualized the results of every submission. Through our inspection, we discovered some common successes and failures.

Since we use Average Precision as the primary metric, participating teams tended to submit a large number of redundant detections with lower confidence scores in order to get higher recalls. Therefore, for every submission, we filter the detections by the threshold that results in the maximum F-score, and visualize the remaining detections on images.

We noticed that the detection performance on digital-born images is generally better than that on natural images. Figure~\ref{fig:task1-examples} (i) shows such an example. The reason is likely to be cleaner background and simpler fonts.

A common mistake we have discovered is failing to detect long text. Figure~\ref{fig:task1-examples} (a), (c), and (f) are examples of this kind. Long text lines are often not fully detected, \emph{i.e.} missing a few characters, or detected in multiple separate pieces. We believe that without further refinement, it is usually hard to produce boxes with large aspect ratios accurately, using popular object detection frameworks such as Faster R-CNN. Some methods also mentioned they used a strategy of connecting multiple segments. But that strategy seems prone to failed connections.

Another common mistake is failing to suppress redundant detections. Such examples are shown in Figure~\ref{fig:task1-examples} (b), (g), and (h). Standard non-maximum suppression struggles to handle these cases, as the redundant detections are much smaller than whole text lines. Smarter suppressing strategy is needed for these cases.

For Task 2, the performance of End-to-End Recognition is dependent on the text localization. Without doubt, worse localization results in worse recognition, like Figure~\ref{fig:task2-examples} (a). For the recognition itself, we observed that perspective distortions harm the performance badly, such as Figure~\ref{fig:task2-examples} (b).

Another common mistake is confusing characters with similar structure. Figure~\ref{fig:task2-examples} (c) shows failure cases of this kind. These Chinese characters are harder to distinguish for their subtle differences.

\section{Conclusion}

We organized the first RCTW-17 competition. A new dataset was collected and released. We also proposed new evaluation protocols that are designed for Chinese text reading. During the challenge, we received a good number of registrations and submissions, which indicates the broad interest on this topic in the community. Through analysis on the results, we shed some light on the challenges and difficulties of this problem.

In the future, we plan to make the challenge a long-term and continuous one. To this end, we plan to set up an online evaluation website where participants can submit, evaluate, and compare their methods at any time.
We will also keep improving the annotations of CTW-12k by correcting mistakes and adding new images.

\section*{Acknowledgment}

The challenge is supported in part by NSFC 61222308.
The authors thank Dr. Fei Yin and Dr. Cheng-Lin Liu for their suggestions.
The authors also thank Zhiyong Liu, Yang Yang, Zhiqiang Zhang, Rui Yu and Xuelei Zhang for their efforts in annotating the data.



\bibliographystyle{IEEEtran}
\bibliography{references}
%

\end{document}